\documentclass[10pt,twocolumn,letterpaper]{article}

\usepackage{cvpr}      

\usepackage{booktabs}
\usepackage[table]{xcolor}
\usepackage{multirow}
\usepackage{colortbl}
\usepackage{tabularx}

\definecolor{darkred}{RGB}{180,40,40}
\newcommand{\gain}[1]{\,\textcolor{green!55!black}{\scriptsize$\blacktriangle$\,#1}}

\definecolor{cvprblue}{rgb}{0.21,0.49,0.74}
\usepackage[pagebackref,breaklinks,colorlinks,allcolors=cvprblue]{hyperref}

\def\confName{CVPR}
\def\confYear{2027}

\title{ReTrace: Cross-Stage Failure Attribution for Self-Improving \\Robotic Manipulation}

\author{
Yue Yu$^{1,3}$ \quad
Bin Zhu$^{2}$ \quad
Junhui Li$^{1}$ \quad
Jiayu Wang$^{1,3}$ \quad
Shijie Zhou$^{1,3}$ \quad
Jingjing Chen$^{3,\dagger}$ \\
$^{1}$College of Computer Science and Artificial Intelligence, Fudan University \\
$^{2}$Singapore Management University \\
$^{3}$Institute of Trustworthy Embodied AI, Fudan University \\
}
\begin{document}
\maketitle

\renewcommand{\thefootnote}{$\dagger$}
\footnotetext{Corresponding Author.}

\begin{abstract}
General-purpose multimodal foundation models are increasingly capable of controlling robots, yet reliably completing complex manipulation tasks remains challenging, particularly those involving cross-stage dependencies. In such tasks, an action may achieve its immediate objective yet cause downstream failures.
Enabling general-purpose agents to learn from failures to complete such cross-stage dependent tasks through autonomous interaction remains challenging. To address this challenge, we introduce ReTrace, a framework for self-improving robotic manipulation through autonomous physical interaction, without demonstrations or model-parameter updates. ReTrace retrospectively traces downstream failures to earlier decisions, proposes targeted corrections, and evaluates them based on changes in execution across attempts. ReTrace progressively consolidates validated corrections into a reusable manipulation skill that guides subsequent execution. We evaluate ReTrace on four real-world manipulation tasks using a Franka Research 3 robot. On the two tasks involving cross-stage dependencies, GPT-6 Astra-low improves from 0\% to 100\%  success on ball pouring and from 0\% to 80\% on plug insertion when equipped with skills acquired through ReTrace. Substantial improvements are also observed for both GPT-6.1 Sol-high and Sol-low, demonstrating the effectiveness of ReTrace.
\end{abstract}
\vspace{-3mm}

\section{Introduction}

Recent advances in general-purpose multimodal foundation models are beginning to reshape robotic manipulation. Frontier foundation models, such as GPT-6 Astra~\cite{gpt6astra}, exhibit strong visual understanding, reasoning, and action-generation capabilities, making it increasingly feasible to use a general-purpose agent as a closed-loop robot controller without task-specific policy training~\cite{zhang2026unexpected, team2026systematically}. Equipped with appropriate harnesses~\cite{chen2026show, yang2026astranomy, tanaka2026taczero, wang2026spatialharness}, these models have demonstrated promising performance on a range of manipulation tasks, including those requiring fine-grained control. However, despite these advances, reliably completing complex manipulation tasks remains challenging, particularly when successful execution depends on decisions made across multiple stages. More importantly, while existing agents can often recover from immediate execution errors through observation and replanning, their ability to learn from failures that originate in earlier stages remains limited.

\begin{figure}[t]
    \centering
    \includegraphics[width=8.1cm]{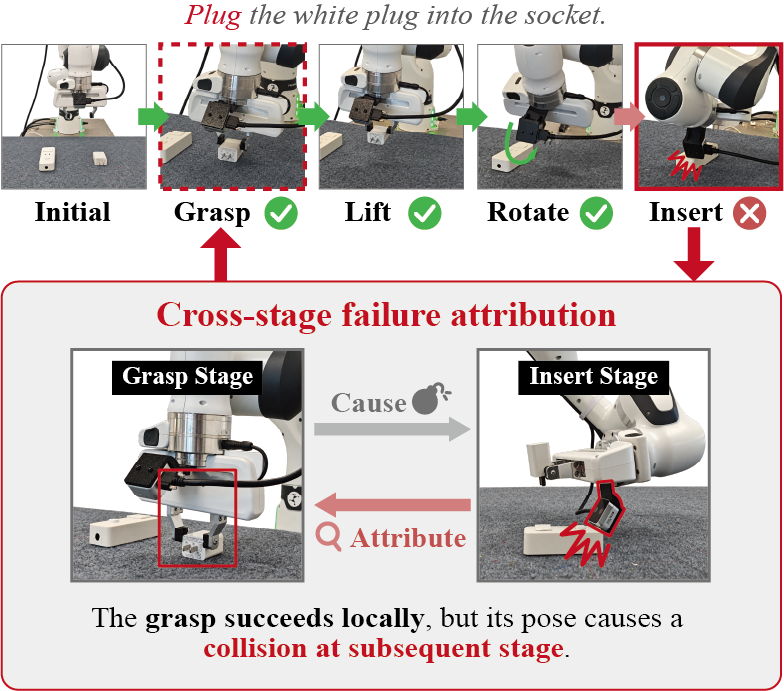}
    \caption{\textbf{Cross-stage dependencies and failure attribution in robotic manipulation.} Top: Plug insertion involves grasping, lifting, rotating, and inserting the plug. Although the grasp succeeds, the resulting relative pose between the gripper and plug leaves insufficient clearance for subsequent insertion, causing a collision. Bottom: Retrospective reasoning traces the insertion collision back to the earlier grasp pose, identifying the decision to revise.}
    \label{fig:dependency}
    \vspace{-3mm}
\end{figure}

\begin{figure}[]
    \centering
    \includegraphics[width=8.2cm]{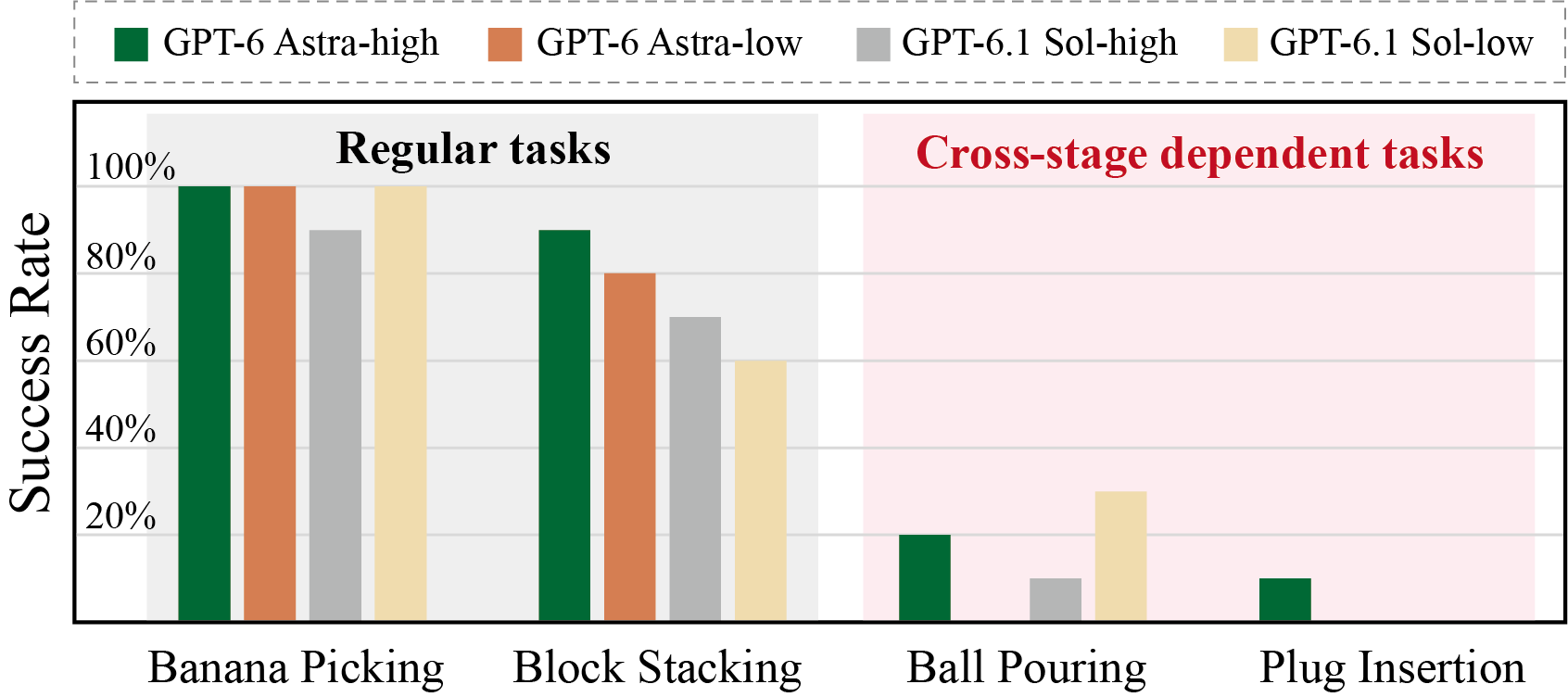}
    \caption{\textbf{Frontier models' zero-shot success rates on regular and cross-stage dependent tasks.} Frontier models perform well on regular tasks but achieve low success rates on cross-stage dependent tasks. The latter remain challenging even for frontier models.}
    \label{fig:tasks_comparison}
    \vspace{-5mm}
\end{figure}

A fundamental challenge arises from \textbf{\emph{cross-stage dependencies}}, where an action may successfully achieve its immediate objective yet create unfavorable conditions that cause downstream failures. Unlike local execution failures, which can often be identified and corrected through immediate feedback, such failures may only become apparent several stages after the responsible decision was made.
As illustrated in Figure~\ref{fig:dependency}, a robot may successfully grasp, lift, and rotate a plug, but an unfavorable initial grasp pose can leave insufficient clearance for the final insertion, resulting in a collision. Although the failure manifests during insertion, adjusting the insertion trajectory alone may not resolve the underlying problem. Instead, the robot must reconsider its earlier grasp configuration. For such tasks, even frontier models that perform well on regular manipulation tasks exhibit substantially lower zero-shot success rates, as shown in Figure~\ref{fig:tasks_comparison}.
These results raise a central question: Can general-purpose agents learn from failures to complete such cross-stage dependent tasks through autonomous interaction? Yet, this learning process involves two main difficulties. First, downstream failures may originate from earlier decisions, making it difficult to identify what and how to correct. Second, the effects of a correction may only become apparent during downstream execution. This delayed feedback makes it difficult to assess whether a correction is effective. Incorrect judgments may cause the agent to overlook potentially effective solutions. 
However, conventional closed-loop replanning and generic reflection \cite{huang2023innermonologue, liu2023reflect} do not explicitly address tasks involving cross-stage dependencies and the two problems they pose.

To address this challenge, we introduce \textbf{ReTrace}, a framework for self-improving robotic manipulation through cross-stage failure attribution and progressive correction, which enables a general-purpose multimodal foundation model to acquire task-specific manipulation knowledge through iterative real-world interaction, without additional demonstrations or model-parameter updates. ReTrace comprises an \textit{Executor}, an \textit{Attributor}, and an evolving \textit{Experience Memory}. The Executor performs closed-loop planning and action execution based on current observations and accumulated experience. After each attempt, the Attributor retrospectively analyzes the execution trajectory, tracing observed failures to potentially responsible earlier decisions and proposing targeted corrections. Importantly, ReTrace evaluates proposed corrections based on the changes they induce across attempts rather than overall task success. Corrections that demonstrate progress can be retained and progressively refined, even when the overall task remains unsuccessful, while ineffective corrections can be discarded. The resulting evidence is consolidated into reusable manipulation skills that guide future execution. Through this attribution-driven improvement process, ReTrace transforms interaction failures into actionable knowledge, enabling the agent to progressively address dependencies that are difficult to resolve through immediate feedback alone.


We evaluate ReTrace on four real-world manipulation tasks using a Franka Research 3 robot, including two regular tasks and two tasks with cross-stage dependencies. Experimental results demonstrate substantial improvements in task success, particularly on cross-stage dependency tasks. Specifically, GPT-6 Astra-low achieves 100\% success rate on ball pouring and 80\% on plug insertion with the acquired skills, compared with 0\% on both tasks without them. Substantial improvements are also observed for both GPT-6.1 Sol-high and Sol-low. In addition, the learned skills can also be reused across model and reasoning configurations, while reducing inference-time decision counts in most evaluated settings.

Our main contributions are as follows:
\begin{itemize}
    \item We identify learning from failures in tasks involving \textit{cross-stage dependencies} as a key challenge for robotic manipulation, where downstream failures may originate from earlier decisions, requiring the agent to trace downstream failures back to earlier decisions and propose targeted corrections.
    
    \item We propose ReTrace, which combines retrospective failure attribution, targeted correction, cross-attempt evaluation, and experience consolidation to acquire a task-specific manipulation skill through autonomous physical interaction, without demonstrations or parameter updates.

    \item We conduct real-world experiments on four manipulation tasks, demonstrating substantial improvements in success rates, particularly on tasks involving strong \textit{cross-stage dependencies}, as well as improved skill learning efficiency through autonomous interaction.

\end{itemize}

\section{Related Work}

\subsection{General-purpose models for robotic manipulation.}
Robotic manipulation has traditionally relied on policies trained on robot trajectories. Vision-language-action (VLA) models such as RT-2~\cite{zitkovich2023rt2}, OpenVLA~\cite{kim2024openvla}, $\pi_0$~\cite{black2024pi0}, and $\pi_{0.5}$~\cite{intelligence2025pi05} follow this paradigm by adapting pretrained representations to robot action prediction. However, recent advances in general-purpose language and multimodal models~\cite{gpt6astra, anthropic2026claudeopus55, kavukcuoglu2026gemini4argon} have opened a new possibility: leveraging their reasoning and decision-making capabilities for robotic manipulation without fine-tuning. Methods such as Code as Policies~\cite{liang2023code}, CaP-X~\cite{fu2026capx} and ASPIRE~\cite{lu2026aspire} leverage general-purpose models to synthesize executable robot control programs from task instructions. These programs coordinate perception, planning, and control tools to execute robotic manipulation tasks. More recently, studies~\cite{chen2026show, zhang2026unexpected, yang2026astranomy, li2026robo, cheng2026context} have shown that appropriate harness designs enable general-purpose models to directly serve as robot policies for closed-loop control. These methods enable frontier agents to control robots through structured action interfaces and execution feedback. Building on these closed-loop control capabilities, ReTrace focuses on enabling general-purpose agents to tackle challenging tasks involving cross-stage dependencies.

\begin{figure*}[t]
    \centering
    \includegraphics[width=17.5cm]{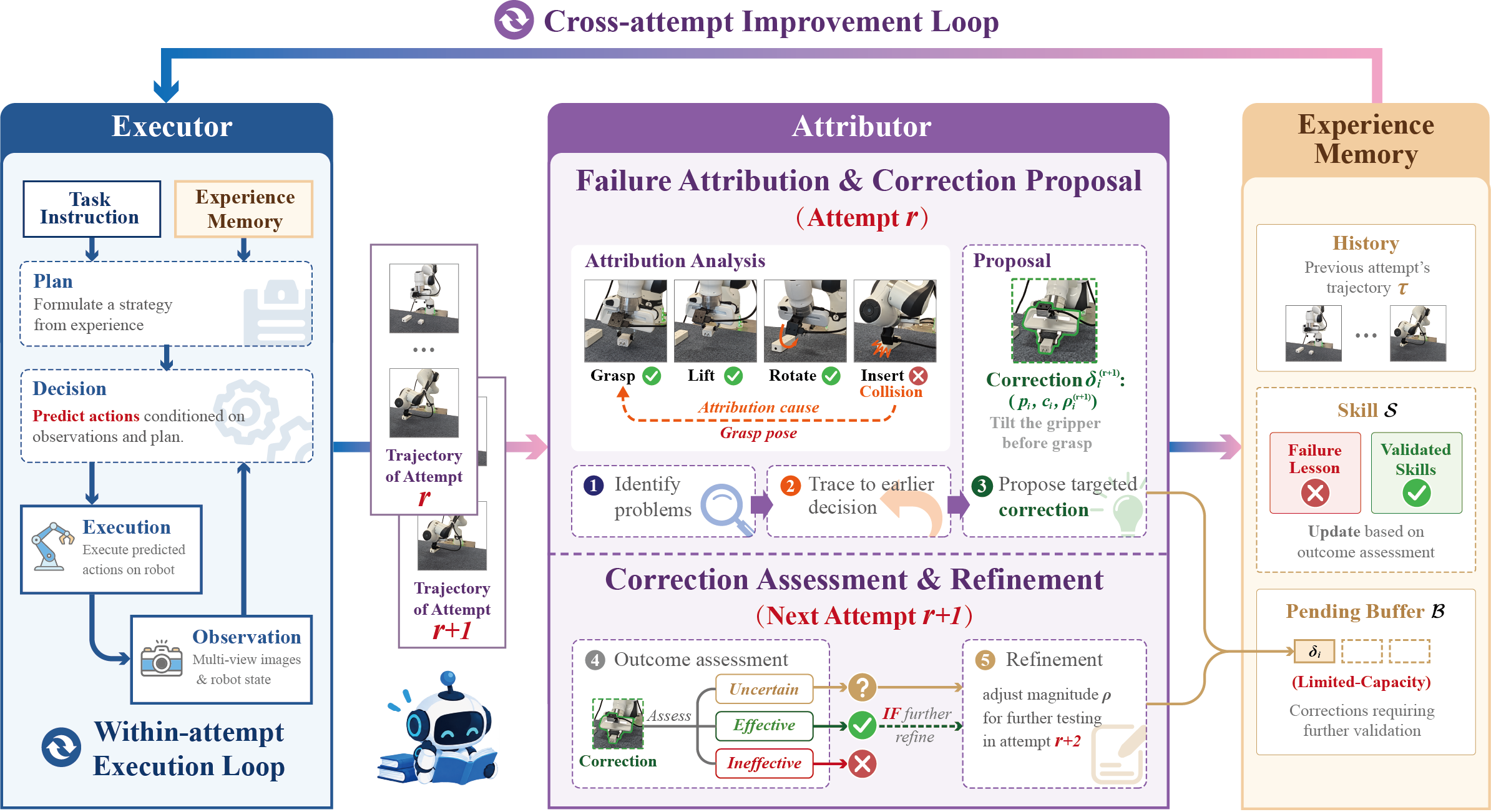}
    \caption{\textbf{Overview of the framework.} The Executor plans and executes robot actions using current observations and accumulated experience. Across attempts, the Attributor diagnoses problems, proposes corrections, and refines them based on execution feedback. The Experience Memory stores skill, pending corrections, and execution history to guide subsequent attempts.}
    \label{framework}
\end{figure*}

\subsection{Experience-driven Self-improvement for Embodied Agents.}
Experience-driven self-improvement aims to turn experience from past attempts into reusable knowledge for future decisions. Prior studies leverage mechanisms such as reflection, memory, and reusable skills to improve agent performance~\cite{madaan2023selfrefine,shinn2023reflexion,wang2024voyager,zhao2024expel, zhang2026ace}. In robotics, existing works also explore how execution feedback and accumulated experience can improve performance. Inner Monologue~\cite{huang2023innermonologue} and REFLECT~\cite{liu2023reflect} leverage execution feedback for plan revision and failure correction. Harness VLA~\cite{zhang2026harnessvla} leverages execution memory to guide agentic planning and coordinate robot control primitives. Its open-source framework also supports closed-loop robot control with general-purpose multimodal models. Zeva~\cite{chen2026zeva} leverages causal memory from physical interactions to guide a frozen robot policy. Other studies ~\cite{fu2026capx,lu2026aspire,zhang2026playful,wang2026harnesspai} explore control program refinement and reusable skill acquisition through interaction with the environment. These advances motivate our exploration of how agents can learn from failures involving cross-stage dependencies in robotic manipulation.

\section{Method}

We propose \textbf{ReTrace}, an embodied self-improvement framework, to enable general-purpose models to develop task-specific capabilities from scratch through interactions with the physical world. To address issues arising from cross-stage dependencies, ReTrace introduces an attribution-driven improvement mechanism that identifies the causes of observed problems and proposes targeted corrections. Across attempts, ReTrace progressively accumulates knowledge and consolidates supported findings into a reusable skill without any additional demonstrations, task-specific prior knowledge, or model-parameter updates.

\subsection{Two-Agent Framework}

Given a task instruction $I$, the current observation $o_t=(V_t,s_t)$, and a general-purpose model $G$, our goal is to enable $G$ to learn task-specific skills $\mathcal{S}$ through iterative interactions with the physical world. Here, $V_t$ denotes visual observations, and $s_t$ denotes the robot state at time $t$. As illustrated in Figure~\ref{framework}, ReTrace is a two-agent framework comprising three core components: an \textbf{Executor}, an \textbf{Attributor}, and an \textbf{Experience Memory}. Both agents are instantiated from $G$. The Executor handles online planning and action prediction, while the Attributor performs retrospective reasoning, proposes corrections, and updates memory. The Experience Memory maintains gained experience, pending corrections, and execution history of the previous attempt to support the Executor's inference and the Attributor's analysis. Together, these components support two loops: a \textbf{within-attempt loop} that interacts with the real world, and a \textbf{cross-attempt loop} that enables progressive improvement through retrospective reasoning over attempts and their outcomes.

\textbf{Within-attempt execution loop.} The Executor is provided with the robot control interface specification. Each attempt begins with the robot in a fixed initial state. Specifically, the Executor first formulates a plan based on the task instruction and initial observations. If this is not the first attempt, the Executor also incorporates Experience Memory into the plan. Guided by this plan, the Executor carries out the task on the real-world robot through multiple steps. At each step, the Executor predicts an action chunk from the available context and current observation $o_t$. After the chunk is executed on the real-world robot, the Executor uses the resulting observations and execution feedback to assess the outcome and inform its next decision. This observation--decision--execution--feedback cycle continues until the attempt ends.

\textbf{Cross-attempt improvement loop.} When the $r$-th attempt ends, its execution trajectory $\tau_r$ is passed to the Attributor. When $r>1$, the evolving memory also retains the previous attempt's trajectory $\tau_{r-1}$ as history for comparison. The Attributor analyzes these trajectories to diagnose problems and propose targeted corrections with the proposed attribution-driven improvement mechanism. It also updates the Experience Memory based on the current attempt's execution outcomes and formulates a strategy for the next attempt. This loop continues until the task is successfully completed.

\textbf{Evolving Experience Memory.} The Experience Memory comprises skill $\mathcal{S}$, a pending buffer $\mathcal{B}$, and execution history. The skill retains effective practices and failure lessons supported by previous interactions, while the pending buffer stores target corrections to be tested in the next attempt. The execution history retains the previous attempt’s trajectory for comparison. This organization distinguishes supported knowledge from pending corrections while preserving the execution context needed to evaluate them.

\subsection{Attribution-Driven Progressive Improvement}
\label{sec:attribution}

Complex manipulation tasks often involve cross-stage dependencies: an action may achieve its local objective while rendering subsequent operations infeasible. Addressing such failures requires tracing the problem back to earlier actions that may have caused it and making targeted improvements. ReTrace therefore introduces Attribution-Driven Progressive Improvement, which connects failure attribution, correction proposal and refinement, and skill consolidation across exploration attempts to guide the agent toward task completion.

\textbf{Failure Attribution and Correction Proposal.}
After the $r$-th attempt, the Attributor reviews the execution trajectory $\tau_r$ to identify where progress stalled and why. Using visual observations, measured robot states, action commands, and execution feedback from trajectory, it analyzes whether the problem arose from a local issue at the point where progress stalled or from an earlier stage that achieved its immediate goal but hindered subsequent execution. As shown in Figure~\ref{framework}, the robot successfully grasps, lifts, and rotates the plug, but the chosen grasp pose creates a collision risk during subsequent insertion. The Attributor attributes this collision risk to the earlier grasp pose, which leaves insufficient clearance for insertion.

Based on failure attribution, the Attributor proposes targeted corrections, specifying how the Executor should change its behavior to address the identified problems in attempt $r+1$. Each correction is represented as
\begin{equation}
    \delta_i^{(r+1)} = \bigl(p_i, c_i, \rho_i^{(r+1)}\bigr),
\end{equation}
where $p_i$ specifies the problem, $c_i$ specifies what to adjust and how, and $\rho_i^{(r+1)}$ specifies the planned magnitude for the next attempt. Each correction concerns a single factor. Corrections are stored in the pending buffer $\mathcal{B}$ with capacity $K$, where $|\mathcal{B}|\leq K$. The limited capacity helps control the number of factors changed at once, making it easier for the Attributor to assess whether each correction is effective. New proposals are admitted only when capacity is available and must be compatible with other corrections currently in the buffer.

\begin{figure*}[t]
    \centering
    \includegraphics[width=15.5cm]{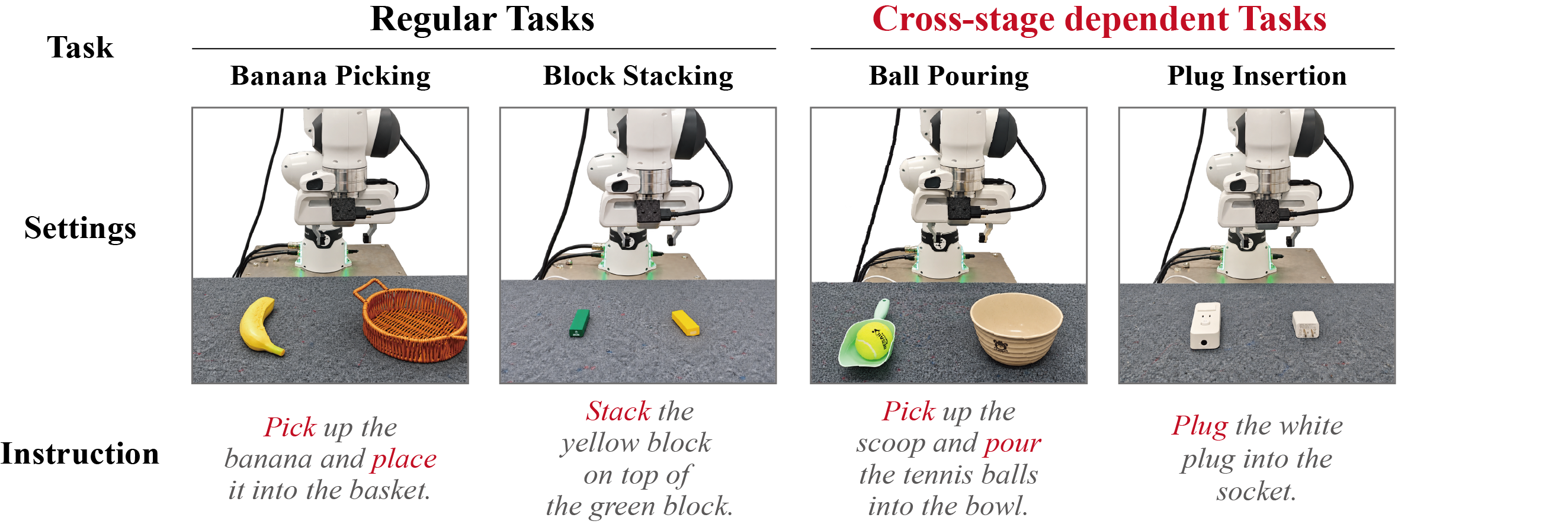}
    \caption{\textbf{Real-world task settings.} From left to right: banana picking, block stacking, ball pouring, and plug insertion. The latter two tasks require earlier grasping decisions to support later operations.}
    \label{fig:setup}
\end{figure*}

\textbf{Correction Assessment and Refinement.}
For corrections applied during attempt $r$, the Attributor compares current attempt’s trajectory $\tau_r$ with the previous trajectory $\tau_{r-1}$ at corresponding stages. It checks whether the intended change actually occurred and evaluates its effect on the associated problem $p_i$. Each correction is classified as \emph{effective, ineffective, or uncertain}. A correction is effective if it makes progress in resolving the associated problem, even without fully resolving it or completing the task. It is ineffective when evidence shows that it does not alleviate the problem. When the evidence is inconclusive, the correction is classified as uncertain. 

Based on this assessment, the Attributor determines whether each correction should be retained in the pending buffer $\mathcal{B}$ for another attempt. Corrections are removed from the buffer if their associated problems have been resolved or they are deemed ineffective. Corrections are retained if they have made progress but may benefit from further refinement, or if their effects remain uncertain and require further evaluation. For retained corrections, the Attributor refines them for testing in subsequent attempts. Specifically, $p_i$ and $c_i$ remain fixed, while the Attributor adjusts the magnitude $\rho_i^{(r+1)}$ based on the current results.

\begin{table*}[t]
    \centering
    \caption{Task success rates (\%). The GPT-6 Astra-high baseline is shaded gray. The highest value in each column is shown in bold, including ties. Green triangles indicate absolute improvements in percentage points over the corresponding model configurations without skills. + ReTrace denotes models equipped with skills learned through ReTrace.}
    \label{tab:results}
    \small
    \setlength{\tabcolsep}{4pt}
    \renewcommand{\arraystretch}{1.15}
    \begin{tabularx}{\textwidth}{@{}l*{6}{>{\centering\arraybackslash}X}@{}}
        \toprule
        \multirow{2}{*}{Model}
        & \multicolumn{3}{c}{\textbf{Regular Tasks}}
        & \multicolumn{3}{c}{\textcolor{darkred}{\textbf{Cross-stage Dependent Tasks}}} \\
        \cmidrule(lr){2-4} \cmidrule(lr){5-7}
        & Banana Picking
        & Block Stacking
        & Average
        & Ball Pouring
        & Plug Insertion
        & Average \\
        \midrule
        \rowcolor{gray!15}
        GPT-6 Astra-high
        & \textbf{100\%} & 90\% & 95\%
        & 20\% & 10\% & 15\% \\
        \midrule
        GPT-6 Astra-low
        & \textbf{100\%} & 80\% & 90\%
        & 0\% & 0\% & 0\% \\
        \ \ + \textbf{ReTrace}
        & \textbf{100\%}
        & \textbf{100\%} \gain{20}
        & \textbf{100\%} \gain{10}
        & \textbf{100\%} \gain{100}
        & 80\% \gain{80}
        & \textbf{90\%} \gain{90} \\
        \midrule
        GPT-6.1 Sol-high
        & 90\% & 70\% & 80\%
        & 10\% & 0\% & 5\% \\
        \ \ + \textbf{ReTrace}
        & \textbf{100\%} \gain{10}
        & \textbf{100\%} \gain{30}
        & \textbf{100\%} \gain{20}
        & 70\% \gain{60}
        & \textbf{90\%} \gain{90}
        & 80\% \gain{75} \\
        \midrule
        GPT-6.1 Sol-low
        & \textbf{100\%} & 60\% & 80\%
        & 30\% & 0\% & 15\% \\
        \ \ + \textbf{ReTrace}
        & \textbf{100\%}
        & \textbf{100\%} \gain{40}
        & \textbf{100\%} \gain{20}
        & 80\% \gain{50}
        & 60\% \gain{60}
        & 70\% \gain{55} \\
        \bottomrule
    \end{tabularx}
\end{table*}

\textbf{Skill Consolidation.}
The Attributor updates the skill $\mathcal{S}$ based on its assessment of the corrections. Corrections assessed as effective are recorded as reusable practices, together with their effects and applicability conditions, while confirmed ineffective corrections are recorded as failure lessons. An effective correction can be recorded in $\mathcal{S}$ while remaining in the pending buffer $\mathcal{B}$ for further refinement. Other effective behaviors observed during execution can also be incorporated into the skill. The updated skill guides the Executor's planning and execution in the next attempt. Across attempts, ReTrace accumulates effective practices and failure lessons, helping the Executor improve its behavior and learn how to complete the task.

\section{Experiments}

\subsection{Experimental Setup}

We conduct real-world experiments on a Franka Research 3 robot across four manipulation tasks: banana picking, block stacking, ball pouring, and plug insertion. The experimental settings are shown in Figure~\ref{fig:setup}. Banana picking and block stacking serve as regular tasks without strong cross-stage dependencies. Conversely, ball pouring and plug insertion involve cross-stage dependencies. In the ball pouring task, the scoop must be grasped at its balance point to maintain stability during subsequent lifting and pouring. In plug insertion, the grasp pose must allow subsequent rotation and insertion without collision. For robot control, a general-purpose multimodal agent is employed to directly predicts a chunk of robot actions at each decision step. More implementation details are provided in Appendix~\ref{appendix_implementation}.

We use GPT-6 Astra with high reasoning effort (GPT-6 Astra-high) to learn task-specific skills through ReTrace. The acquired skills are then evaluated across different models and reasoning-effort settings by comparing execution with and without them. For each evaluated configuration, we conduct 10 trials per task and report the success rate. The per-trial decision budget is set to 25 for all tasks except plug insertion, for which the budget is 35 because it involves fine-grained manipulation. Furthermore, we compare the cross-attempt learning efficiency of ReTrace with Harness VLA \cite{zhang2026harnessvla} and two ablated variants. We also report the performance of GPT-6 Astra-high without learned skills as a baseline.
 

\subsection{Main Results}
\textbf{Effectiveness of Learned Skills.} 
We use ReTrace with GPT-6 Astra-high to learn task-specific skills for all four tasks. The effectiveness of these skills is then evaluated using GPT-6 Astra-low, GPT-6.1 Sol-high, and GPT-6.1 Sol-low~\cite{openai2026gpt61sol}. The results are summarized in Table~\ref{tab:results}. For tasks without strong cross-stage dependencies, banana picking is relatively simple, with baseline success rates already reaching 90–100\%. With skills learned by ReTrace, all three configurations achieve 100\% success. Block stacking shows greater improvements, with success rates increasing from 80\% to 100\% for GPT-6 Astra-low, from 70\% to 100\% for GPT-6.1 Sol-high, and from 60\% to 100\% for GPT-6.1 Sol-low.
 
For cross-stage dependent tasks, ReTrace’s learned skills lead to substantial improvements. Without learned skills, all three settings achieve low success rates on ball pouring (0–30\%) and 0\% on plug insertion. GPT-6 Astra-high, which is used for skill learning, also achieves only 20\% on ball pouring and 10\% on plug insertion. When equipped with the learned skills, GPT-6 Astra-low improves from 0\% to 100\% success on ball pouring and from 0\% to 80\% on plug insertion. GPT-6.1 Sol-high and Sol-low also show substantial improvements on both tasks. 

These results highlight the challenges and significance of cross-stage dependent tasks, even for frontier models that perform well on regular manipulation tasks. Furthermore, the improvements across different models and reasoning-effort settings demonstrate the effectiveness and transferability of the task-specific skills acquired through the proposed ReTrace.



\begin{figure}[t]
    \centering
    \includegraphics[width=8.1cm]{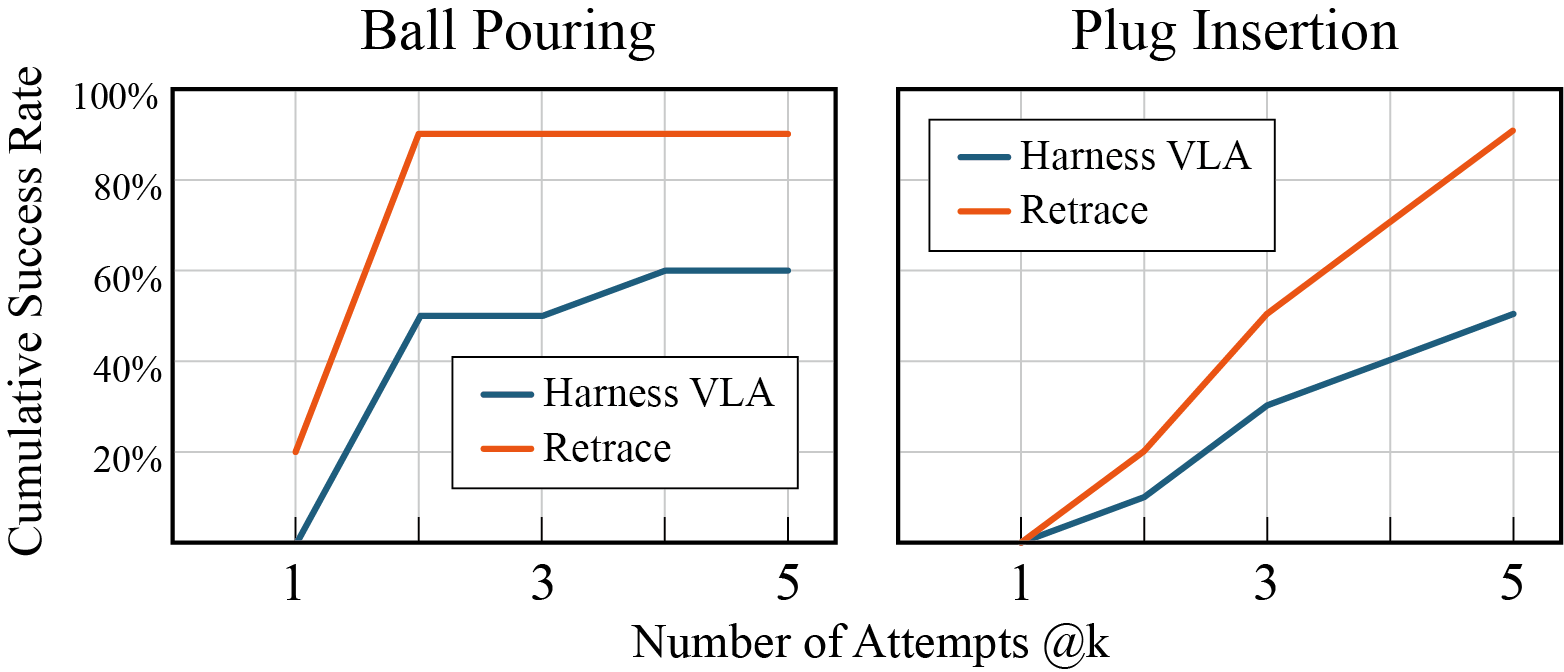}
    \caption{Cross-attempt learning efficiency of ReTrace and Harness VLA on Ball Pouring (left) and Plug Insertion (right). Both methods employ GPT-6 Astra with high reasoning effort and are evaluated over 10 independent exploration runs per task.}
    \label{fig:learning_efficiency}
\end{figure}

\begin{table}[t]
\centering
\small
\caption{Cross-attempt learning efficiency. GPT-6 Astra-high@1 reports
the single-attempt success rate of the base model. Both Harness VLA and ReTrace employ GPT-6 Astra with high reasoning effort. @k report cumulative success rates (\%) within the first $k$
attempts. Average reports the mean success rate across the two tasks.}
\label{tab:learning_efficiency}
\begin{tabular*}{\columnwidth}{@{\extracolsep{\fill}}lccc@{}}
\toprule
Setting & Ball Pour. & Plug Ins. & Average \\
\midrule
GPT-6 Astra-high@1 & 20\% & 10\% & 15\% \\
\midrule
Harness VLA@1 & 0\% & 0\% & 0\% \\
Harness VLA@3 & 50\% & 30\% & 40\% \\
Harness VLA@5 & 60\% & 50\% & 55\% \\
\midrule
ReTrace@1
& 20\%
& 0\%
& 10\% \\
ReTrace@3
& \textbf{90\%} \gain{40}
& 50\% \gain{20}
& 70\% \gain{30} \\
ReTrace@5
& \textbf{90\%} \gain{30}
& \textbf{90\%} \gain{40}
& \textbf{90\%} \gain{35} \\
\bottomrule
\end{tabular*}
\end{table}

\textbf{Cross-Attempt Learning Efficiency.}
We further evaluate cross-attempt learning efficiency against Harness VLA~\cite{zhang2026harnessvla}. Its open-source agentic framework supports memory-based iterative exploration and direct robot control with general-purpose multimodal agent. We adapt this framework to Franka Research 3 for comparison. Both methods employ GPT-6 Astra-high for exploration. We compare the two methods on two cross-stage dependent tasks, Plug Insertion and Ball Pouring. In this experiment, each exploration run allows the agent to learn how to complete the task through up to five iterative attempts. For each task, we conduct 10 independent exploration runs per method to measure the cumulative success rate (Success@k), defined as the percentage of exploration runs that succeed within the first $k$ attempts.

\begin{table*}[t]
    \centering
    \caption{Ablation study of \textit{Failure Attribution} and \textit{Progressive Improvement} on cross-attempt learning efficiency. All settings use GPT-6 Astra with high reasoning effort. \textit{+ two agent} denotes the two-agent framework without
    Failure Attribution or Progressive Improvement. The full ReTrace setting combines the two-agent framework with both components. GPT-6 Astra-high@1 reports the single-attempt success rate of the base model without learned skills. For all other settings, @k reports the cumulative success rate (\%) within the first $k$ attempts. Mean reports the average success rate across the two tasks. The highest value in each column is shown in bold, including ties.}
    \label{tab:learning_efficiency_ablation}
    \small
    \setlength{\tabcolsep}{4pt}
    \renewcommand{\arraystretch}{1.15}
    \begin{tabularx}{\textwidth}{@{}l*{3}{>{\centering\arraybackslash}X}@{}}
        \toprule
        Setting & Ball Pouring & Plug Insertion & Mean \\
        \midrule

        GPT-6 Astra-high@1
        & 20\% & 10\% & 15\% \\

        \midrule

        \textit{+ two agent}@1
        & 20\% & 0\% & 10\% \\

        \textit{+ two agent}@3
        & 40\% & 40\% & 40\% \\

        \textit{+ two agent}@5
        & 50\% & 60\% & 55\% \\

        \midrule

        \textit{+ two agent + Failure Attribution}@1
        & 10\% & 0\% & 5\% \\

        \textit{+ two agent + Failure Attribution}@3
        & 60\% & 60\% & 60\% \\

        \textit{+ two agent + Failure Attribution}@5
        & 60\% & 70\% & 65\% \\

        \midrule

        \textit{+ two agent + Failure Attribution + Progressive Improvement (ReTrace)}@1
        & 20\% & 0\% & 10\% \\

        \textit{+ two agent + Failure Attribution + Progressive Improvement (ReTrace)}@3
        & \textbf{90\%} & 50\% & 70\% \\

        \textit{+ two agent + Failure Attribution + Progressive Improvement (ReTrace)}@5
        & \textbf{90\%} & \textbf{90\%} & \textbf{90\%} \\

        \bottomrule
    \end{tabularx}
\end{table*}

As shown in Table~\ref{tab:learning_efficiency} and Figure~\ref{fig:learning_efficiency}, both methods exhibit low first-attempt success rates, suggesting that these tasks remain challenging. As exploration continues, our ReTrace achieves average cumulative success rates of 70\% and 90\% within three and five attempts, respectively, compared with 40\% and 55\% for Harness VLA. Notably, ReTrace surpasses Harness VLA's five-attempt performance within only three attempts, demonstrating that ReTrace learns more efficiently from real-world interactions. Our observations suggest that the proposed Progressive Improvement mechanism contributes to this advantage. During real-world exploration, a corrective strategy proposed by the agent may not be effective at first if it is not applied to the appropriate degree. ReTrace can retain such a potential strategy and progressively refine how it should be applied in subsequent attempts. In contrast, Harness VLA tends to try a different strategy after an unsuccessful attempt, which may cause it to abandon a potentially effective solution and limit its cumulative success rate. These results show that our method helps the agent learn more efficiently and effectively from real-world interactions with a limited budget.

\subsection{Ablation Study}

To validate the effectiveness of the proposed Attribution-Driven Progressive Improvement mechanism, we conduct an ablation study on two cross-stage dependent tasks. Specifically, we decompose the mechanism into two components: Failure Attribution and Progressive Improvement. We compare two ablated variants with the full ReTrace. The first variant, denoted by \textit{+ two agent}, removes both components while retaining the two-agent structure. Under this setting, the Attributor still reviews the current attempt's execution trajectory to update the skill and generate suggestions for the Executor's next attempt. The second variant, denoted by \textit{+ two agent + Failure Attribution}, incorporates Failure Attribution but removes Progressive Improvement. In this variant, the Attributor performs cross-stage failure attribution and generates targeted suggestions for the next attempt. In particular, these suggestions are not formulated as structured corrections, managed with a pending buffer, or subjected to the Correction Assessment and Refinement procedure. The full ReTrace framework is denoted by \textit{+ two agent + Failure Attribution + Progressive Improvement (ReTrace)}. We adopt the same experimental protocol as in the cross-attempt learning efficiency evaluation (Sec.~4.2), using GPT-6 Astra-high for skill learning.

\begin{figure}[]
    \centering
    \includegraphics[width=8.2cm]{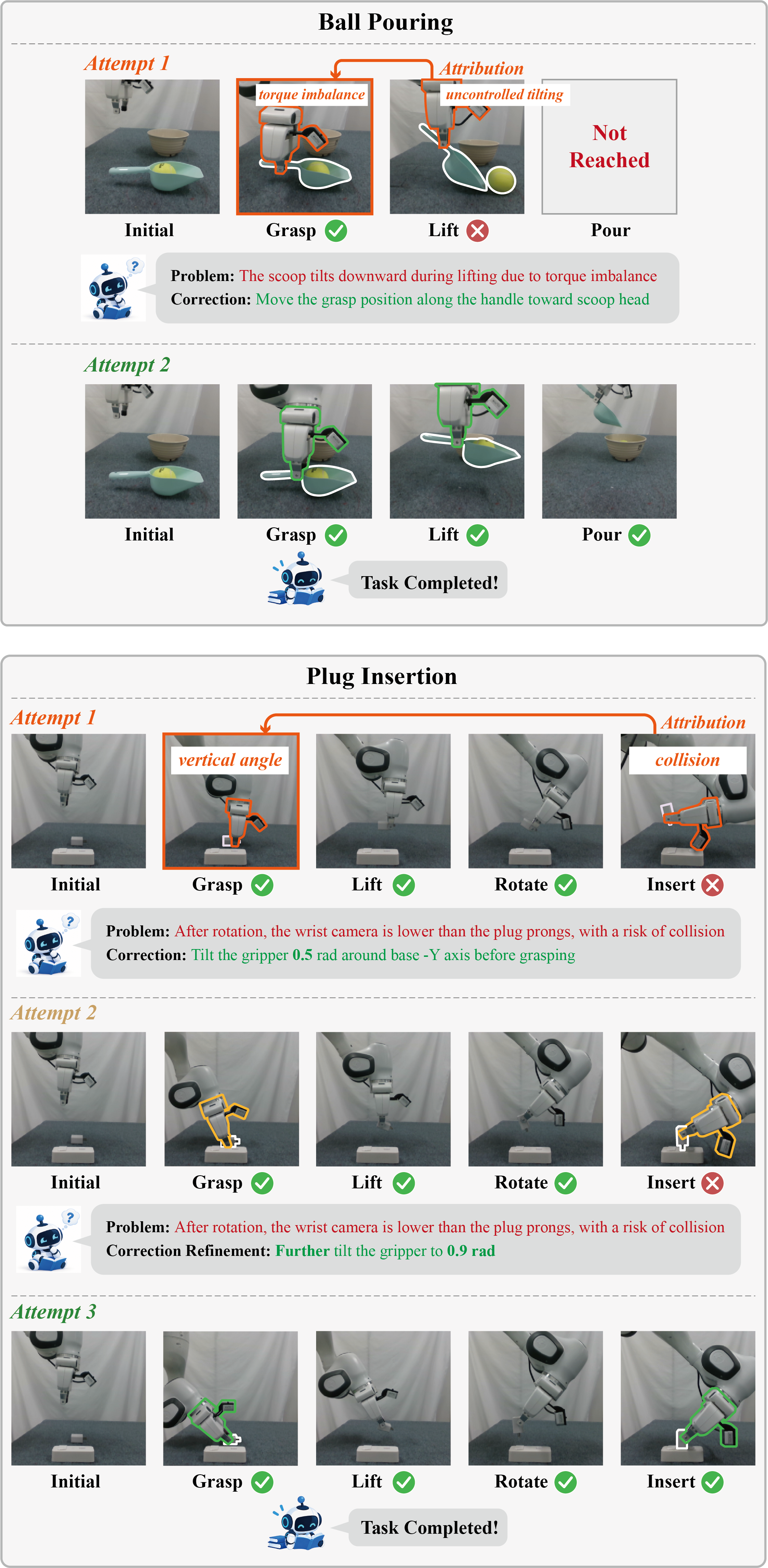}
    \caption{\textbf{Learning process through cross-stage failure attribution and correction refinement.} \textbf{Top:} Ball pouring succeeds in the second attempt by shifting the grasp position to prevent scoop instability during lifting.  \textbf{Bottom:} Plug insertion improves over three attempts by attributing the collision risk to an unfavorable grasp pose and progressively refining the gripper tilt before grasping.
    }
    \label{fig:learning}
\end{figure}

\begin{figure*}[t]
    \centering
    \includegraphics[width=17.5cm]{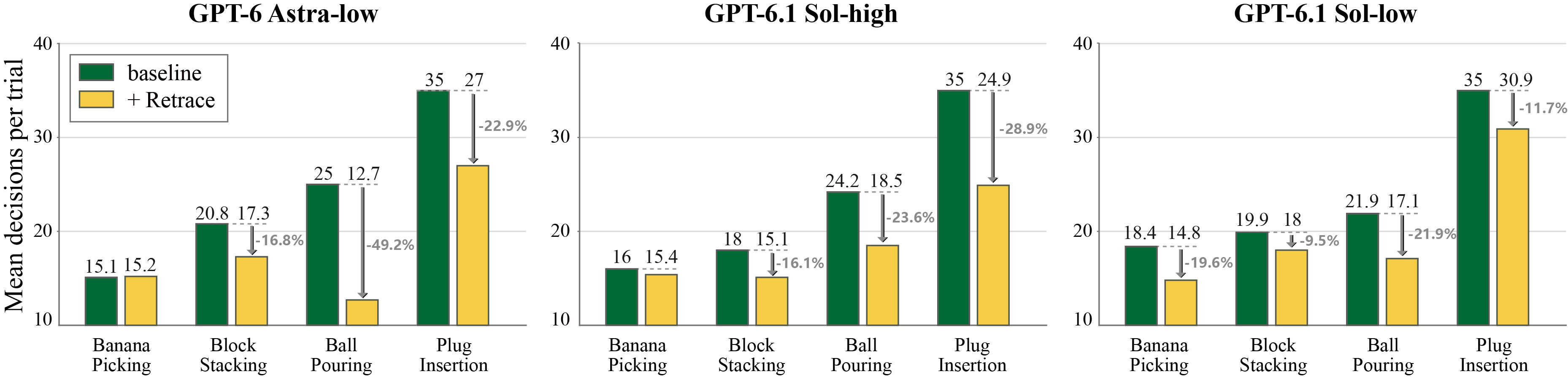}
    \caption{\textbf{Inference-time decision counts with and without ReTrace.} Bars show the mean over 10 trials per configuration. The per-trial decision budget is 35 for plug insertion and 25 for each of the other tasks. Unsuccessful trials are assigned the full task-specific decision budget.}
    \label{fig:decision_count}
\end{figure*}

As shown in Table~\ref{tab:learning_efficiency_ablation}, adding Failure Attribution to the two-agent framework improves Success@3 and Success@5 on both tasks. The mean Success@3 and Success@5 increase from 40\% to 60\% and from 55\% to 65\%, respectively. These results suggest that explicit failure attribution facilitates skill learning on cross-stage dependent tasks.  Incorporating Progressive Improvement further increases the mean Success@5 to 90\%. Our observations during the experiments suggest that Progressive Improvement contributes to this gain in two ways. First, Progressive Improvement helps retain potentially effective corrections and refine them across attempts. In contrast, both ablated variants tend to abandon potentially effective strategies after a single failed attempt and switch to alternatives, which may limit further improvement. On Plug Insertion, the cumulative success rate of full ReTrace increases from 50\% within three attempts to 90\% within five attempts. Over the same attempt budgets, the {\it + two agent + Failure Attribution} variant only improves from 60\% to 70\%. This pattern is consistent with the goal of progressively refining corrections. Second, structured corrections specify the problem to address, the proposed adjustment, and its magnitude, providing more concrete guidance for subsequent attempts. We observed that the Executor of the {\it + two agent + Failure Attribution} variant did not consistently follow the Attributor's suggestions during execution. In contrast, the full ReTrace framework enables the Executor to better implement the proposed corrections. This helps explain the increase in Success@3 on Ball Pouring from 60\% for the {\it + two agent + Failure Attribution} variant to 90\% for full ReTrace. Overall, these ablation results demonstrate the effectiveness of the proposed Attribution-Driven Progressive Improvement mechanism.

\subsection{Visualizing the Learning Process}

Figure~\ref{fig:learning} illustrates how ReTrace learns from cross-stage failures through retrospective attribution and correction refinement in ball pouring and plug insertion. For ball pouring, in Attempt 1, the Executor successfully grasps the scoop handle, but the scoop becomes unstable and tilts downward during lifting due to torque imbalance. Continued lifting causes the ball to roll out of the scoop, preventing further progress. Then, the Attributor traces this torque imbalance problem back to the earlier grasp position and proposes moving the grasp position toward the scoop head. In Attempt 2, this correction stabilizes the scoop during lifting, allowing the robot to complete the pouring task.

For plug insertion, in Attempt 1, the Executor successfully grasps and lifts the plug, but the grasp pose causes the wrist camera to be lower than the plug prongs after rotation, creating a collision risk. The Attributor traces this failure back to the earlier grasp pose and proposes tilting the gripper by 0.5 rad before grasping. In Attempt 2, the correction improves the grasp configuration but remains insufficient to avoid the collision risk. Rather than discarding this correction, the Attributor retains it and increases the tilt to 0.9 rad for further exploration. In Attempt 3, the refined grasp pose enables successful insertion. Together, these examples demonstrate how ReTrace leverages retrospective reasoning to address failures arising from cross-stage dependencies and improve subsequent attempts through targeted corrections.

\subsection{Inference-Time Decision Cost Analysis}
We measure inference-time decision cost by the number of model calls during task execution. Each call generates a chunk of robot actions and counts as one decision. For each task and model configuration, we compare execution with and without the acquired skills and report the mean decision count over 10 trials per condition. If the task is not completed within the decision budget, we use the full budget as the decision count. As illustrated in Figure~\ref{fig:decision_count}, acquired skills reduce mean decision counts in 11 of the 12 evaluated settings, with the magnitude of the reduction varying across tasks and model configurations.

For simple tasks such as banana picking, GPT-6 Astra-low and GPT-6.1 Sol-high show limited improvement in decision efficiency with the acquired skills. Sol-low benefits more, with its mean decision count decreasing from 18.4 to 14.8. These results suggest that skills offer limited additional gains when the model already executes the task efficiently. More substantial reductions in mean decision counts are observed on the cross-stage dependent tasks. Together with the improved success rates in Table~\ref{tab:results}, these results demonstrate the benefits of the acquired skills for completing challenging tasks.


\section{Conclusion}
Tasks involving cross-stage dependencies are challenging because locally successful actions can cause downstream failures. These tasks offer a valuable setting for studying how agents learn to complete complex tasks through autonomous interaction. We introduce ReTrace, which traces such failures to earlier decisions, evaluates targeted corrections based on execution changes across attempts, and consolidates validated corrections into reusable skills without demonstrations or model-parameter updates. Experiments on four real-world tasks demonstrate improved success rates, particularly on cross-stage dependent tasks, and more efficient skill learning.

{
    \small
    \bibliographystyle{ieeenat_fullname}
    \bibliography{main}

@String(AAAI = {AAAI})

@article{team2026systematically,
  title={Systematically exploring the capabilities of GPT-6 Astra as embodied policies},
  author={Team, Galbot and Chen, Xuchuan and Cheng, Xiaoqian and Deng, Yu and Ding, Lihe and Dong, Shaocong and Gao, Xiangjun and Jia, Haozhe and Li, Zekai and Li, Zhoujian and others},
  journal={arXiv preprint arXiv:2609.38537},
  year={2026}
}

@inproceedings{zitkovich2023rt2,
  title = {{RT-2: Vision-Language-Action Models Transfer Web Knowledge to Robotic Control}},
  author = {Zitkovich, Brianna and Yu, Tianhe and Xu, Sichun and Xu, Peng and Xiao, Ted and Xia, Fei and Wu, Jialin and Wohlhart, Paul and Welker, Stefan and Wahid, Ayzaan and Vuong, Quan and Vanhoucke, Vincent and Tran, Huong and Soricut, Radu and Singh, Anikait and Singh, Jaspiar and Sermanet, Pierre and Sanketi, Pannag R. and Salazar, Grecia and Ryoo, Michael S. and Reymann, Krista and Rao, Kanishka and Pertsch, Karl and Mordatch, Igor and Michalewski, Henryk and Lu, Yao and Levine, Sergey and Lee, Lisa and Lee, Tsang-Wei Edward and Leal, Isabel and Kuang, Yuheng and Kalashnikov, Dmitry and Julian, Ryan and Joshi, Nikhil J. and Irpan, Alex and Ichter, Brian and Hsu, Jasmine and Herzog, Alexander and Hausman, Karol and Gopalakrishnan, Keerthana and Fu, Chuyuan and Florence, Pete and Finn, Chelsea and Dubey, Kumar Avinava and Driess, Danny and Ding, Tianli and Choromanski, Krzysztof Marcin and Chen, Xi and Chebotar, Yevgen and Carbajal, Justice and Brown, Noah and Brohan, Anthony and Arenas, Montserrat Gonzalez and Han, Kehang},
  booktitle = {Conference on Robot Learning},
  series = {Proceedings of Machine Learning Research},
  volume = {229},
  pages = {2165--2183},
  year = {2023},
  url = {https://proceedings.mlr.press/v229/zitkovich23a.html},
}

@article{black2024pi0,
  title = {{$\pi_0$: A Vision-Language-Action Flow Model for General Robot Control}},
  author = {Black, Kevin and Brown, Noah and Driess, Danny and Esmail, Adnan and Equi, Michael and Finn, Chelsea and Fusai, Niccolo and Groom, Lachy and Hausman, Karol and Ichter, Brian and Jakubczak, Szymon and Jones, Tim and Ke, Liyiming and Levine, Sergey and Li-Bell, Adrian and Mothukuri, Mohith and Nair, Suraj and Pertsch, Karl and Shi, Lucy Xiaoyang and Tanner, James and Vuong, Quan and Walling, Anna and Wang, Haohuan and Zhilinsky, Ury},
  journal = {arXiv preprint arXiv:2410.24164},
  year = {2024},
  url = {https://arxiv.org/abs/2410.24164},
}

@article{intelligence2025pi05,
  title = {{$\pi_{0.5}$: a Vision-Language-Action Model with Open-World Generalization}},
  author = {{Physical Intelligence} and Black, Kevin and Brown, Noah and Darpinian, James and Dhabalia, Karan and Driess, Danny and Esmail, Adnan and Equi, Michael and Finn, Chelsea and Fusai, Niccolo and Galliker, Manuel Y. and Ghosh, Dibya and Groom, Lachy and Hausman, Karol and Ichter, Brian and Jakubczak, Szymon and Jones, Tim and Ke, Liyiming and LeBlanc, Devin and Levine, Sergey and Li-Bell, Adrian and Mothukuri, Mohith and Nair, Suraj and Pertsch, Karl and Ren, Allen Z. and Shi, Lucy Xiaoyang and Smith, Laura and Springenberg, Jost Tobias and Stachowicz, Kyle and Tanner, James and Vuong, Quan and Walke, Homer and Walling, Anna and Wang, Haohuan and Yu, Lili and Zhilinsky, Ury},
  journal = {arXiv preprint arXiv:2504.16054},
  year = {2025},
  url = {https://arxiv.org/abs/2504.16054},
}

@inproceedings{madaan2023selfrefine,
  title = {{Self-Refine: Iterative Refinement with Self-Feedback}},
  author = {Madaan, Aman and Tandon, Niket and Gupta, Prakhar and Hallinan, Skyler and Gao, Luyu and Wiegreffe, Sarah and Alon, Uri and Dziri, Nouha and Prabhumoye, Shrimai and Yang, Yiming and Gupta, Shashank and Majumder, Bodhisattwa Prasad and Hermann, Katherine and Welleck, Sean and Yazdanbakhsh, Amir and Clark, Peter},
  booktitle = {Advances in Neural Information Processing Systems},
  volume = {36},
  year = {2023},
  url = {https://arxiv.org/abs/2303.17651},
}

@inproceedings{shinn2023reflexion,
  title = {{Reflexion: Language Agents with Verbal Reinforcement Learning}},
  author = {Shinn, Noah and Cassano, Federico and Berman, Edward and Gopinath, Ashwin and Narasimhan, Karthik and Yao, Shunyu},
  booktitle = {Advances in Neural Information Processing Systems},
  volume = {36},
  year = {2023},
  url = {https://arxiv.org/abs/2303.11366},
}

@article{wang2024voyager,
  title = {{{Voyager}: An open-ended embodied agent with large language models}},
  author = {Wang, Guanzhi and Xie, Yuqi and Jiang, Yunfan and Mandlekar, Ajay and Xiao, Chaowei and Zhu, Yuke and Fan, Linxi and Anandkumar, Anima},
  journal = {Transactions on Machine Learning Research},
  year = {2024},
  url = {https://arxiv.org/abs/2305.16291},
}

@inproceedings{zhang2026ace,
  title = {{Agentic Context Engineering: Evolving Contexts for Self-Improving Language Models}},
  author = {Zhang, Qizheng and Hu, Changran and Upasani, Shubhangi and Ma, Boyuan and Hong, Fenglu and Kamanuru, Vamsidhar and Rainton, Jay and Wu, Chen and Ji, Mengmeng and Li, Hanchen and Thakker, Urmish and Zou, James and Olukotun, Kunle},
  booktitle = {International Conference on Learning Representations},
  year = {2026},
  url = {https://arxiv.org/abs/2510.04618},
}

@article{zhang2026harnessvla,
  title = {{{Harness VLA}: Steering Frozen VLAs into Reliable Manipulation Primitives via Memory-Guided Agents}},
  author = {Zhang, Yixian and Zhang, Huanming and Gao, Feng and Li, Xiao and Liu, Zhihao and Zhu, Chunyang and Qiu, Jiaxing and Yan, Yuchen and Liu, Jiyuan and Tang, Wenhao and Fang, Zhengru and Nie, Yi and Wei, Changxu and Wang, Yu and Ding, Wenbo and Yu, Chao},
  journal = {arXiv preprint arXiv:2607.08448},
  year = {2026},
  url = {https://arxiv.org/abs/2607.08448},
}

@article{chen2026zeva,
  title = {{Zeva: In-Context Causal Learning for Generalizable Embodied Manipulation}},
  author = {Chen, Fu and Ding, Xin and Huang, Bingjia and Li, Xiangyu and Wang, Mingju and He, Jiawei and Li, Kun and Sun, Wei and Liu, Yunxin and Wu, Hao and Cao, Ting},
  journal = {arXiv preprint arXiv:2608.30880},
  year = {2026},
  url = {https://arxiv.org/abs/2608.30880},
}

@inproceedings{huang2023innermonologue,
  title = {{Inner Monologue: Embodied Reasoning through Planning with Language Models}},
  author = {Wenlong Huang and Fei Xia and Ted Xiao and Harris Chan and Jacky Liang and Pete Florence and Andy Zeng and Jonathan Tompson and Igor Mordatch and Yevgen Chebotar and Pierre Sermanet and Tomas Jackson and Noah Brown and Linda Luu and Sergey Levine and Karol Hausman and Brian Ichter},
  booktitle = {Conference on Robot Learning},
  series = {Proceedings of Machine Learning Research},
  volume = {205},
  pages = {1769--1782},
  year = {2023},
  url = {https://proceedings.mlr.press/v205/huang23c.html},
}

@inproceedings{liu2023reflect,
  title = {{REFLECT: Summarizing Robot Experiences for Failure Explanation and Correction}},
  author = {Liu, Zeyi and Bahety, Arpit and Song, Shuran},
  booktitle = {Conference on Robot Learning},
  series = {Proceedings of Machine Learning Research},
  volume = {229},
  pages = {3468--3484},
  year = {2023},
  url = {https://proceedings.mlr.press/v229/liu23g.html},
}

@article{fu2026capx,
  title = {{{CaP-X}: A framework for benchmarking and improving coding agents for robot manipulation}},
  author = {Fu, Letian and Yu, Justin and El-Refai, Karim and Kou, Ethan and Xue, Haoru and Huang, Huang and Xiao, Wenli and Wang, Guanzhi and Niu, Dantong and Li, Fei-Fei and Shi, Guanya and Wu, Jiajun and Sastry, Shankar and Zhu, Yuke and Goldberg, Ken and Fan, Linxi "Jim"},
  journal = {arXiv preprint arXiv:2603.22435},
  year = {2026},
  url = {https://arxiv.org/abs/2603.22435},
}

@article{lu2026aspire,
  title = {{ASPIRE: Agentic /Skills Discovery for Robotics}},
  author = {Lu, Runyu and Wu, Yubo and Kou, Ethan and Fu, Letian and Xiao, Wenli and Mandlekar, Ajay and Xu, Yinzhen and Shi, Guanya and Goldberg, Ken and Chen, Ang and Chowdhury, Mosharaf and Zhu, Yuke and Fan, Linxi "Jim" and Wang, Guanzhi},
  journal = {arXiv preprint arXiv:2607.00272},
  year = {2026},
  url = {https://arxiv.org/abs/2607.00272},
}

@article{zhang2026playful,
  title = {{Playful Agentic Robot Learning}},
  author = {Zhang, Junyi and Ge, Jiaxin and Yoo, Hanjun and Fu, Letian and Yang, Zihan and Liu, Yaowei and Saravanan, Raj and Yin, Shaofeng and Yu, Justin and Niu, Dantong and Wang, Zirui and Herzig, Roei and Goldberg, Ken and Bai, Yutong and Chan, David M. and Stoica, Ion and Kanazawa, Angjoo and Lei, Jiahui and Feng, Haiwen and Darrell, Trevor},
  journal = {arXiv preprint arXiv:2606.19419},
  year = {2026},
  url = {https://arxiv.org/abs/2606.19419},
}

@article{wang2026harnesspai,
  title = {{HarnessPAI: An Evolving Harness for Physical AI}},
  author = {Wang, Xin and Wu, Wenhao and Zhang, Menghao and Wang, Zhi and Shao, Kun and Luan, Jian and Li, Yang and Li, Qing and Gu, Shangding and Zhou, Huichi and Shi, Shuqing and Ni, Fei and Lu, Shuo and Meng, Weicheng and Li, Kang and Wu, Jin and Zhao, Kang and Guo, Shangmin and Li, Gen and Tang, Yongqiang and Zhang, Zhizhong and Xie, Yuan and Qu, Heng},
  journal = {arXiv preprint arXiv:2609.29166},
  year = {2026},
  url = {https://arxiv.org/abs/2609.29166},
}

@misc{gpt6astra,
  title        = {{GPT-6 Astra: A New Generation of Intelligence}},
  author       = {{OpenAI}},
  howpublished = {OpenAI},
  month        = sep,
  year         = {2026},
  url          = {https://openai.com/index/gpt-6-astra/}
}

@misc{anthropic2026claudeopus55,
  author = {{Anthropic}},
  title = {{Claude Opus 5.5}},
  year = {2026},
  url = {https://www.anthropic.com/claude-opus-5-5}
}

@misc{kavukcuoglu2026gemini4argon,
  author = {Kavukcuoglu, Koray},
  title = {{Gemini 4 Argon}: Our Next Era of Frontier Intelligence},
  year = {2026},
  url = {https://blog.google/innovation-and-ai/models-and-research/gemini-models/gemini-4-argon/}
}

@inproceedings{liang2023code,
  title={Code as policies: Language model programs for embodied control},
  author={Liang, Jacky and Huang, Wenlong and Xia, Fei and Xu, Peng and Hausman, Karol and Ichter, Brian and Florence, Pete and Zeng, Andy},
  booktitle={2023 IEEE International conference on robotics and automation (ICRA)},
  pages={9493--9500},
  year={2023},
  organization={IEEE}
}

@article{zhang2026unexpected,
  title={An unexpected robot policy: Early evaluations of gpt-6 astra on robodojo and beyond},
  author={Zhang, Wenbo and Wang, Kaixuan and Ouyang, Yutao and Huang, Xiaoyu and Li, Liyang and Su, Kailun and Jin, Weiyang and Chai, Wenhao and Liang, Haotian and Dou, Zhiyang and others},
  journal={arXiv preprint arXiv:2609.24170},
  year={2026}
}

@article{yang2026astranomy,
  title={Astranomy: An Empirical Study of GPT-6 Astra as a Native Robot Policy},
  author={Yang, Jenai Xuning and Khetan, Ashna and Zhang, Shun},
  journal={NVIDIA Seattle Robotics Lab},
  year={2026}
}

@article{chen2026show,
  title={Show-Harness: Just a VLM agent can play robots},
  author={Chen, Yanzhe and Bai, Zechen and Cao, Zhijun and Zeng, Wenzheng and Lin, Kevin Qinghong and Lin, Yiqi and Liang, Guoqiang and Ma, Kevin Yuchen and Huang, Qiming and Shou, Mike Zheng},
  journal={arXiv preprint arXiv:2609.10522},
  year={2026}
}

@article{tanaka2026taczero,
  title={TacZero: Training-Free Peg Insertion Using a General-Purpose Vision-Language Model with Tactile Feedback},
  author={Tanaka, Kazutoshi},
  journal={arXiv preprint arXiv:2610.07621},
  year={2026}
}

@article{wang2026spatialharness,
  title={SpatialHarness: Test-Time Spatial Scaffolding for Fine Robotic Manipulation},
  author={Wang, Jiayu and Yu, Yue and Zhu, Bin and Yang, Zhiyao and Chen, Jingjing},
  journal={arXiv preprint arXiv:2610.12457},
  year={2026}
}

@article{li2026robo,
  title={Robo-Harness K1: Harnessing robot-use agents via perception augmentation},
  author={Li, Zexi and Zhang, Yehang and Li, Wenqian and Huang, Haojian and Wang, Chenxu and Deng, Shiyuan and Wei, Yangkai and Zhang, Tianyi and Xie, Binghui and Zhou, Bohan and others},
  journal={arXiv preprint arXiv:2609.29389},
  year={2026}
}

@article{cheng2026context,
  title={In-context robot learning with VLM agents},
  author={Cheng, Dongzhou and Yi, Taoran and Fang, Ye and Zhang, Xingwu and Feng, Fan and Li, Yixuan and Zhuang, Gengxiong and Wang, Rongze and Yang, Shuai and Song, Wei and others},
  journal={arXiv preprint arXiv:2609.19138},
  year={2026}
}

@inproceedings{zhao2024expel,
  title={Expel: Llm agents are experiential learners},
  author={Zhao, Andrew and Huang, Daniel and Xu, Quentin and Lin, Matthieu and Liu, Yong-Jin and Huang, Gao},
  booktitle={Proceedings of the AAAI Conference on Artificial Intelligence},
  volume={38},
  number={17},
  pages={19632--19642},
  year={2024}
}

@article{kim2024openvla,
  title={Openvla: An open-source vision-language-action model},
  author={Kim, Moo Jin and Pertsch, Karl and Karamcheti, Siddharth and Xiao, Ted and Balakrishna, Ashwin and Nair, Suraj and Rafailov, Rafael and Foster, Ethan and Lam, Grace and Sanketi, Pannag and others},
  journal={arXiv preprint arXiv:2406.09246},
  year={2024}
}

@misc{openai2026gpt61sol,
  author       = {{OpenAI}},
  title        = {Introducing GPT-6.1 Sol},
  year         = {2026},
  month        = sep,
  howpublished = {\url{https://openai.com/index/introducing-gpt-6-1-sol/}}
}
}


\clearpage
\appendix
\begin{center}
    {\Large\bfseries Appendix}
\end{center}

\setcounter{figure}{7}
\setcounter{table}{3}

\section{More Implementation Details}
\label{appendix_implementation}

\subsection{Robot Control}

In this section, we provide more details on how the agent controls the robot. Before execution, the general-purpose multimodal agent receives a control interface specification defining the action format, the meaning of each action component, the coordinate frame, and the motion limits of the Franka Research 3 (FR3) arm. Each action consists of seven components. The first six define an {\it end-effector delta-pose command}, comprising a three-dimensional translation vector and a three-dimensional rotation vector, both expressed in the robot base frame. The seventh component commands the gripper to open or close. A robot controller executes the predicted commands sequentially at a nominal rate of 10 Hz. At each control step, it applies fixed action scales of 0.05 m and 0.2 rad to the translation and rotation vectors, respectively, to compute the target end-effector pose. At each decision step, the multimodal agent directly predicts an action chunk from the task instruction, images from the wrist and external cameras, and the current robot state. The agent completes the manipulation task through iterative action prediction and execution.

\textbf{Skill learning with ReTrace.} During skill learning, the ReTrace Executor predicts a fixed-length chunk of 10 actions at each decision step, corresponding to approximately one second of robot motion. Each exploration attempt has a predefined decision budget. To ensure safe and efficient exploration, the Executor can also terminate an attempt early based on safety constraints and task-specific failure criteria.

\textbf{Inference-time execution.} In our experiments, we compare model performance with and without the skills learned through ReTrace. When the learned skill is employed, the model receives it as additional context alongside the task instruction and the control interface specification. Without the learned skill, the model receives the same task instruction and control interface specification. In both settings, the model can autonomously choose to predict a chunk of 10, 20, or 30 actions at each decision step. Each action chunk is counted as one decision. Execution continues until the task is completed or the task-specific decision budget is reached, unless a failure condition is met.

\subsection{Additional Experimental Details}

\noindent\textbf{Camera Setup.}
All experiments use a common set of three cameras: one side-view camera, one front-view camera, and one wrist-mounted camera. In particular, plug insertion requires fine-grained manipulation, so we use an additional camera positioned close to the socket to provide a detailed view of the insertion. For each task, the same camera setup is used across all experiments, model configurations, and methods.

\noindent\textbf{Success and Failure Criteria.}
Across all experiments, a trial is considered successful only if it meets the following criteria for the corresponding task.

\begin{itemize}
    \item \textbf{Banana Picking:} The banana is entirely inside the basket, with no part extending outside it.

    \item \textbf{Block Stacking:} The yellow block is stably stacked on top of the green block and does not touch the table.

        \item \textbf{Ball Pouring:} The ball is stably placed in the bowl, and the bowl does not tip over during execution.

    \item \textbf{Plug Insertion:} The plug's pins are fully inserted into the socket, rather than hovering above it.
\end{itemize}

Across all tasks, a trial is considered a failure if execution is terminated because a commanded action exceeds the  FR3 robot arm's joint limits. For the ball pouring task, the ball falling off the scoop accidentally is an additional failure condition. A trial is immediately marked as failed when any failure condition is met.

\section{Additional Experimental Results}

\begin{table}[b]
    \centering
    \caption{Task success rates (\%) in the OOD settings. Each task success rate is computed over 10 trials. Ball Pour. and Plug Ins. denote Ball Pouring and Plug Insertion, respectively. Average reports the mean success rate across the two tasks. The highest value in each column is shown in bold. Green triangles indicate absolute improvements in percentage points over the corresponding model configurations without skills. + ReTrace denotes agents equipped with skills learned through ReTrace in the standard settings.}
    \label{tab:ood_generalization}
    \small
    \setlength{\tabcolsep}{4pt}
    \renewcommand{\arraystretch}{1.15}
    \begin{tabularx}{\columnwidth}{@{}l*{3}{>{\centering\arraybackslash}X}@{}}
        \toprule
        \multirow{2}{*}{Model}
        & \multicolumn{3}{c}{\footnotesize\textcolor{darkred}{\textbf{Cross-stage Dependent Tasks (OOD Settings)}}} \\
        \cmidrule(lr){2-4}
        & Ball Pour.
        & Plug Ins.
        & Average \\
        \midrule
        GPT-6 Astra-low
        & 20\% & 0\% & 10\% \\
        \ \ + \textbf{ReTrace}
        & \textbf{80\%} \gain{60}
        & \textbf{60\%} \gain{60}
        & \textbf{70\%} \gain{60} \\
        \midrule
        GPT-6.1 Sol-high
        & 20\% & 0\% & 10\% \\
        \ \ + \textbf{ReTrace}
        & 60\% \gain{40}
        & 50\% \gain{50}
        & 55\% \gain{45} \\
        \midrule
        GPT-6.1 Sol-low
        & 0\% & 0\% & 0\% \\
        \ \ + \textbf{ReTrace}
        & 70\% \gain{70}
        & 30\% \gain{30}
        & 50\% \gain{50} \\
        \bottomrule
    \end{tabularx}
\end{table}

\begin{figure*}[t]
    \centering
    \includegraphics[width=13.5cm]{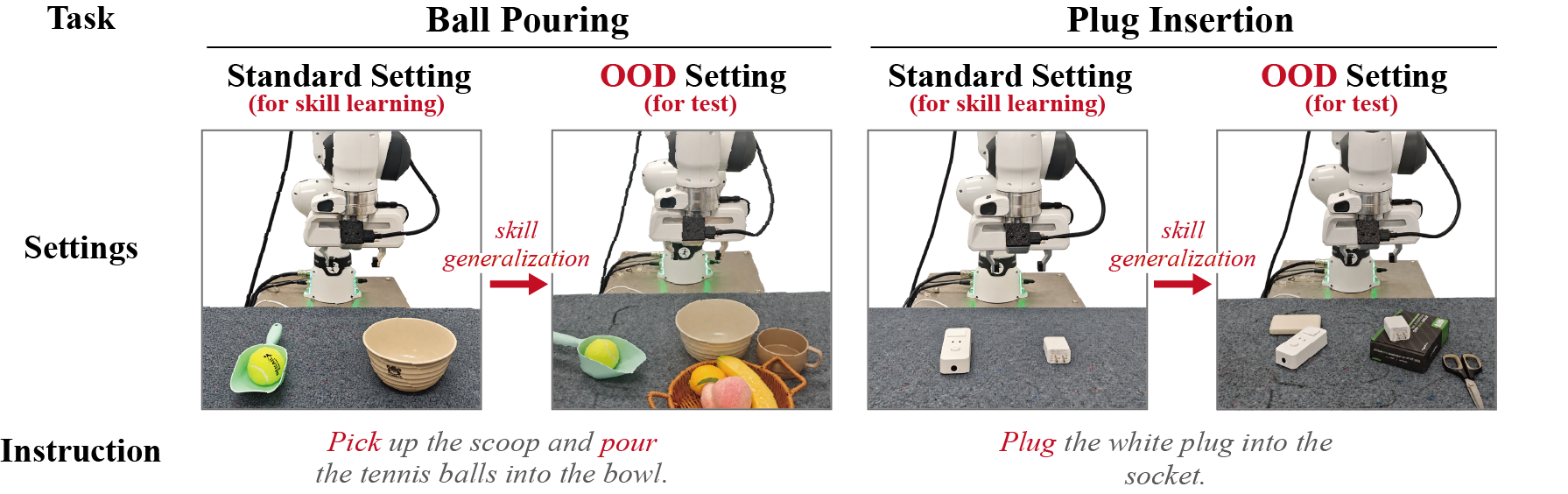}
    \caption{\textbf{OOD settings.} Skills learned through ReTrace in the standard settings are evaluated in the OOD settings.}
    \label{fig:ood_settings}
\end{figure*}

\begin{figure*}[h]
    \centering
    \includegraphics[width=13.5cm]{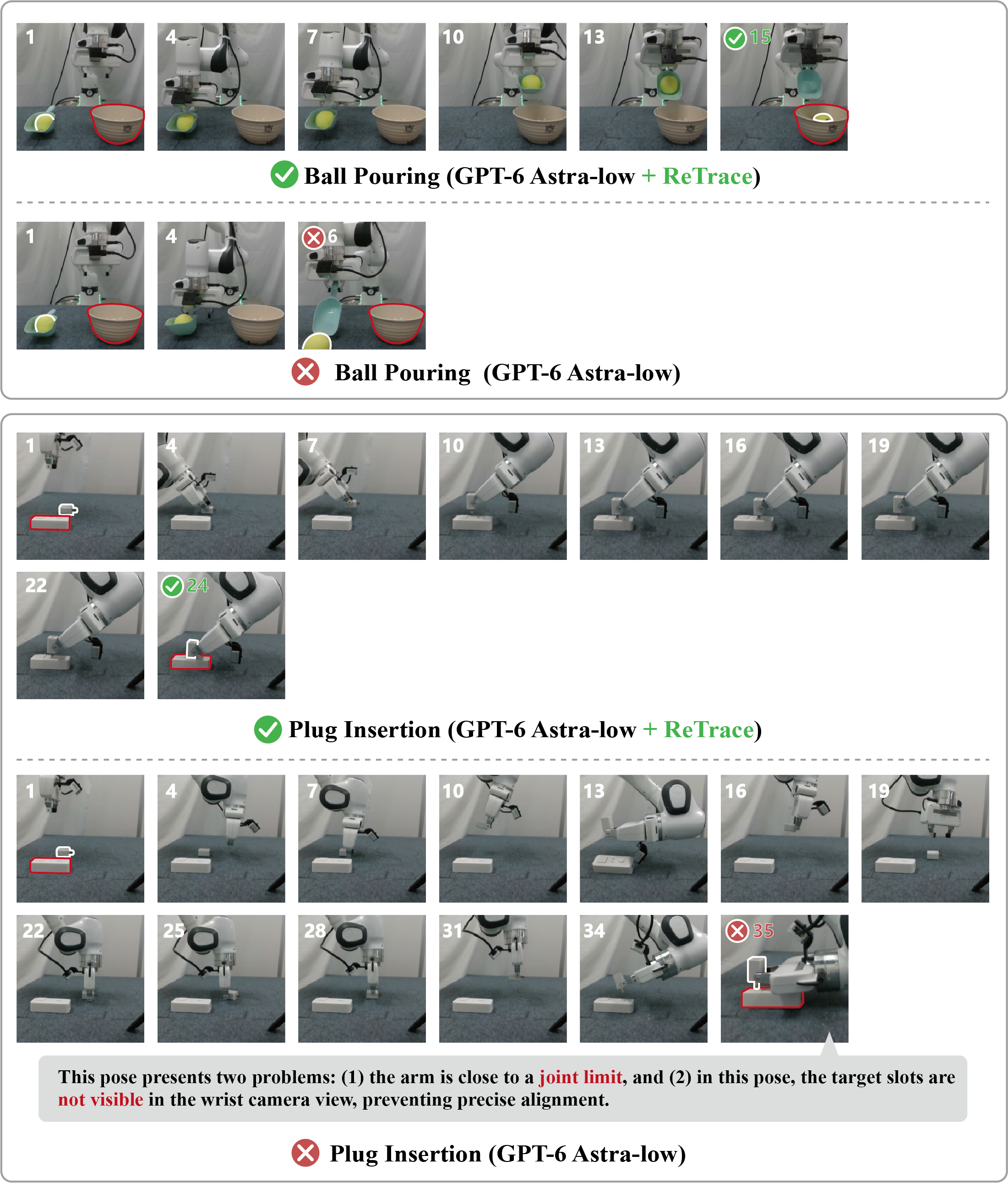}
    \caption{\textbf{Task execution.} GPT-6 Astra-low executes tasks with and without skills learned by ReTrace. Numbers indicate decision steps.}
    \label{fig:visual_std}
\end{figure*}

\begin{figure*}[t]
    \centering
    \includegraphics[width=13.5cm]{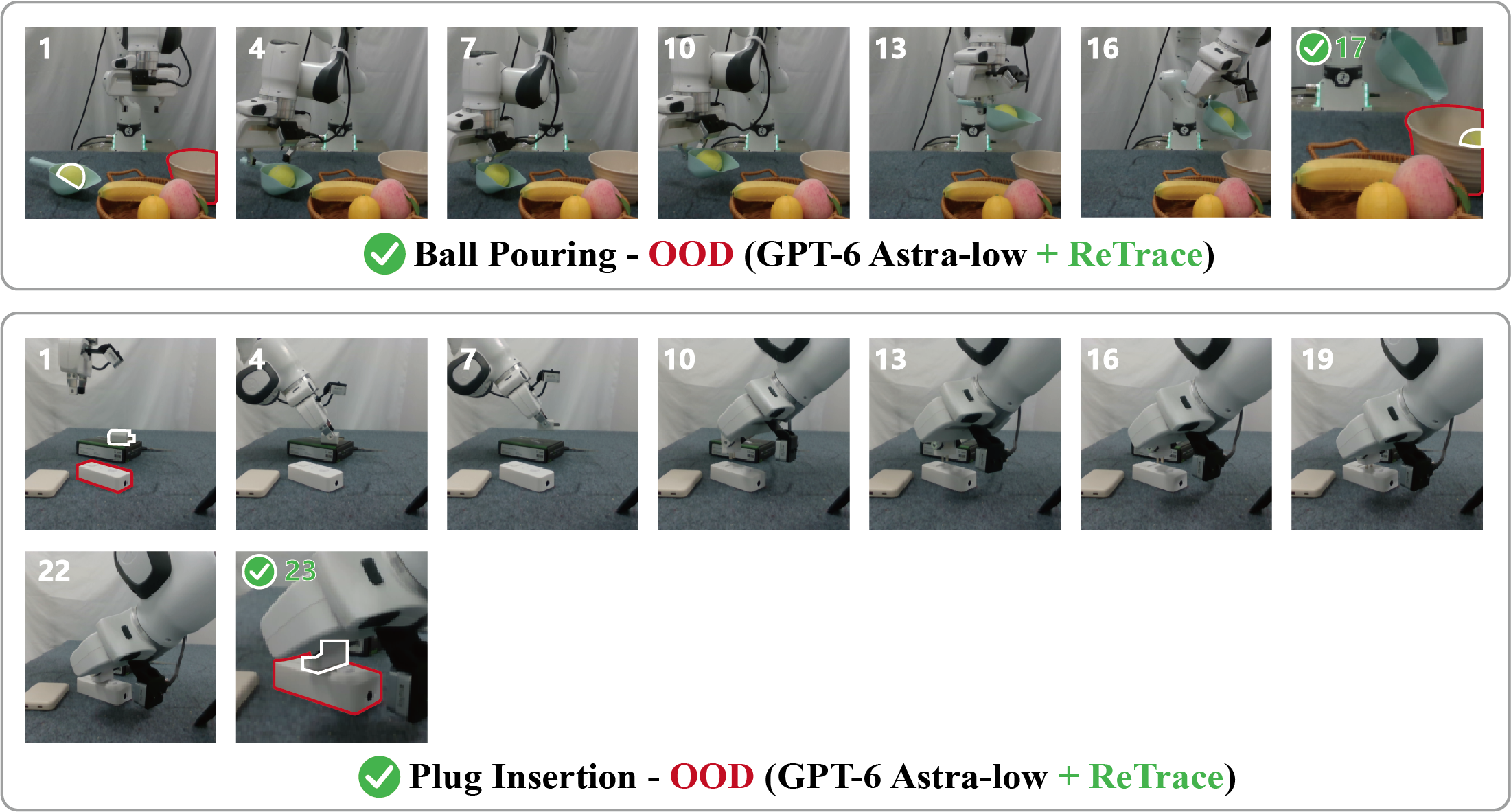}
    \caption{\textbf{Task execution in the OOD settings.} GPT-6 Astra-low completes both tasks using skills learned in the standard settings, demonstrating their generalization to new scenes. Numbers indicate decision steps.}
    \label{fig:visual_ood}
\end{figure*}

\subsection{Generalization of Learned Skills}
To evaluate the generalization of skills learned by ReTrace across scenes, we construct more out-of-distribution (OOD) settings for the two cross-stage dependent tasks by changing object positions and orientations while introducing additional distractors, as shown in Figure~\ref{fig:ood_settings}. Task-specific skills are acquired through ReTrace in the standard settings. These skills are then evaluated in the OOD settings. Specifically, we evaluate GPT-6 Astra-low~\cite{gpt6astra}, GPT-6.1 Sol-high, and GPT-6.1 Sol-low~\cite{openai2026gpt61sol} both with and without the learned skills. For each model configuration, we conduct 10 trials per task under each condition and report the success rate.

As shown in Table~\ref{tab:ood_generalization}, the learned skills still substantially improve task success rates in the OOD settings across all evaluated model configurations on both tasks. For Plug Insertion, the learned skills increase success rates from 0\% to 60\%, 50\%, and 30\% for GPT-6 Astra-low, GPT-6.1 Sol-high, and GPT-6.1 Sol-low, respectively. These gains indicate that skills acquired in the standard settings remain useful in the more complex OOD settings. These results support the generalization and robustness of the learned skills in more complex scenes.

\section{Visualization of Task Execution}
To provide a more intuitive illustration of how the learned skills guide task execution, we present frames from execution trajectories in both standard and OOD settings. In Figure~\ref{fig:visual_std}, we show two execution trajectories for each task in the standard settings, both using GPT-6 Astra-low to control the robot: one with the learned skills and one without. For Ball Pouring task, the agent equipped with learned skill keeps the ball in the scoop during lifting and transport. Without skills, the scoop tilts during lifting, causing the ball to fall out and the task to fail. For Plug Insertion, the agent equipped with the learned skills successfully completes the task. The agent without skills also tries alternative strategies after encountering problems. However, it failed to find a successful strategy across multiple runs of the same experiment. In the example shown in the figure, the agent rotates the gripper by 180° before regrasping the plug, changing the plug’s orientation relative to the wrist camera. This avoids a collision, but the resulting pose remains unsuitable for insertion for two reasons. First, the arm is close to a joint limit, limiting further movement. Second, the target slots are not visible in the wrist camera view, preventing precise alignment. Figure~\ref{fig:visual_ood} shows GPT-6 Astra-low executing both tasks in the OOD settings with skills acquired in the standard settings. Guided by the learned skills, the agent successfully completes both tasks despite the scene changes. These examples further illustrate the generalization of the learned skills to more complex scenes.

\end{document}